\documentclass[a4paper,12pt]{article}
\usepackage{epsfig}

\newcommand{\siff}{\hspace{0.3cm}iff\hspace{0.3cm}}
\newcommand{\sfbox}[1]{\hspace{0.2cm}\fbox{#1}\hspace{0.2cm}}

\title{A General, Sound and Efficient Natural Language Parsing
Algorithm based on Syntactic Constraints Propagation\footnote{Jose F. Quesada:
A General, Sound and Efficient Natural Language Parsing
Algorithm based on Syntactic Constraints Propagation. {\em Proceedings
of CAEPIA'97}, M/'alaga, Spain. 775--786}}

\author{Jos\'e F. Quesada\\
CICA (Centro de Inform\'atica Cient\'{\i}fica de Andaluc\'{\i}a)\\
Sevilla, Spain\\
e-mail: josefran@cica.es
}

\date{}

\begin{document}

\maketitle

\abstract{This paper presents a new context-free parsing algorithm based
on a bidirectional strictly horizontal strategy which incorporates
strong top--down predictions (derivations and adjacencies). From a
functional point of view, the parser is able to propagate syntactic
constraints reducing parsing ambiguity. From a computational perspective,
the algorithm includes different techniques aimed at the improvement
of the manipulation and representation of the structures used.
}

\section{Parsing Ambiguity and Parsing Efficiency}
In Formal Language Theory \cite{Aho+Ullman:72,Drobot:89} 
a language is a set, and in Set Theory
an element belongs or not to a set. That is to say, a set (and
therefore a language) is an unambiguous structure.
A grammar may be considered as an intensive definition of a language.
Thus, the notion of grammaticality corresponds to the relation of
membership over a language (set). But a grammar incorporates more
information than the simple report of the elements of the language
(the extensive specification). A grammar defines a structure: the
parse tree or forest.
The distance between grammaticality and grammatical structure is a
first level of ambiguity: grammatical ambiguity.

The next notion to take into account is the process of analysis of 
a string of words with a grammar, that is, the parser 
\cite{Kay:80,Bolc:87,Sikkel+Nijholt:97}. A parser
must be able to determine the relation of grammaticality and to 
obtain the grammatical structure, by mean of a set of operations,
that we will call the parsing structure.
The distance between the grammatical structure and the parsing
structure defines a second level of ambiguity: parsing ambiguity,
usually referred as temporal ambiguity.

Parsing ambiguity depends on two factors: the grammar and the
parsing strategy.
A very important design requirement of natural language parsers
is to eliminate parsing ambiguity, that is, to reduce the work
done by the parser to the amount of grammatical structures
allowed by the grammar. The work presented here is a step more in this 
direction \cite{Earley:70,Kay:80,Tomita:87,Tomita:91,Dowding+al:94,Bunt+Tomita:
96}.

The second goal of this paper is to present a computational model aimed at
the improvement of the efficiency of the algorithm
\cite{Carroll:94,Quesada+Amores:97}. In this sense, our
proposal may be understood as the incorporation of strong top--down
predictions (partial derivations and adjacencies) over a bottom--up
framework.

And the two strategies (bottom--up and top--down) are mixed by a 
mechanism able to propagate syntactic constraints over a bidirectional
model based on a strictly horizontal strategy \cite{Quesada:96}.

Section 2 presents an informal introduction to the problem of parsing
ambiguity with chart parsing \cite{Kay:80}, but similar situations may be 
described
for other strategies like Earley's algorithm \cite{Earley:70}, DCG 
\cite{Pereira+Warren:80}, GLR \cite{Chapman:87,Tomita:91}, etc. Section
3 defines formally the relations that support the mechanism of
bottom--up bidirectional analysis, top--down predictions and constraints
propagation. Section 4 presents in detail the parsing algorithm and finally
Section 5 shows some experimental results.

\section{An Informal Introduction}

Let us consider the following grammar:

{\scriptsize
\begin{verbatim}
        S  -> A1 b
        S  -> A2 c
        A1 -> a
        A1 -> a A1
        A2 -> a
        A2 -> a A2
\end{verbatim}
}

\noindent and the string of words {\tt a a a b}.
Figure 1 shows the arcs genereted by a bidirectional chart parser in a 
first stage where we have created only the arcs with at least one
pre--terminal symbol. Each arc has been identified by a number, and
indicates the symbol that the arc will generate and the expected
symbol (only for active arcs).

\begin{center}
\epsfxsize=8cm
\epsfbox{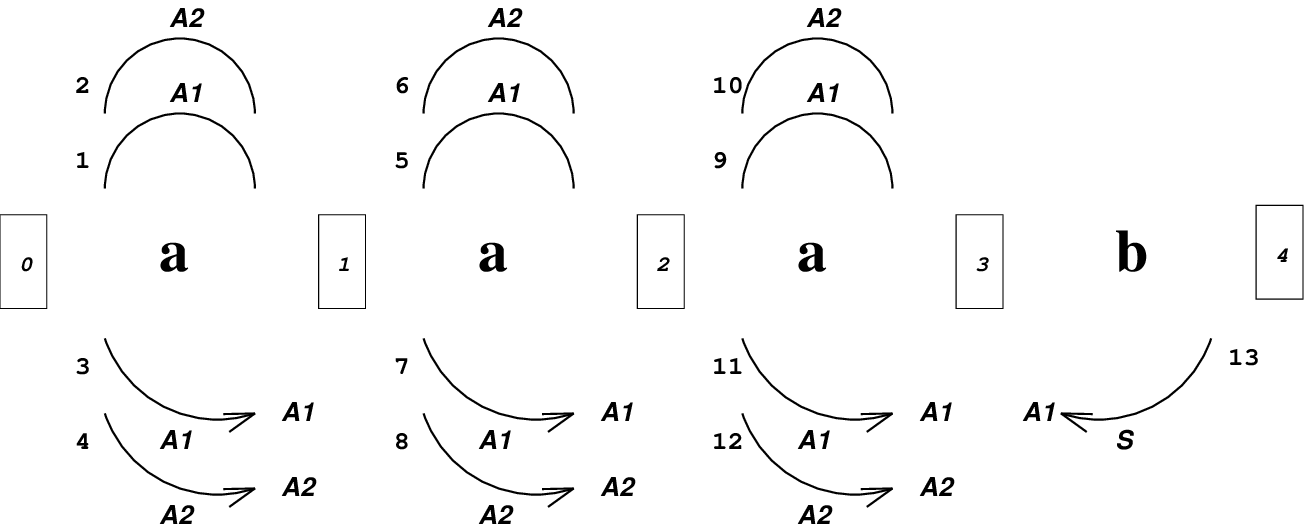}
\end{center}
\begin{center}
\small {\bf Figure 1}
\end{center}

Let us consider what happens at position \fbox{3}. There exists an
obvious relation between arcs \verb!13! and \verb!9!, but arcs \verb!10!,
\verb!11! and \verb!12! don't have a correspondent link at this position. 
Figure 2 shows the parsing state once we have deleted these three arcs.

\begin{center}
\epsfxsize=8cm
\epsfbox{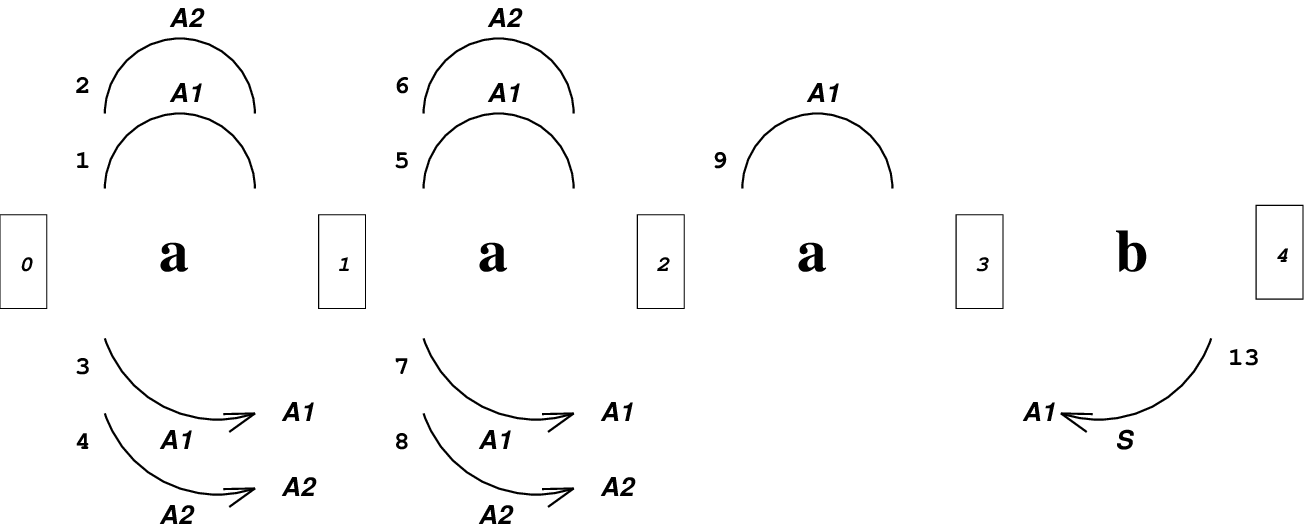}
\end{center}
\begin{center}
\small {\bf Figure 2}
\end{center}
At position \fbox{2} there exists a relation between arcs \verb!7! and 
\verb!9!, and arcs \verb!5!, \verb!6! and \verb!8! may be deleted. Once
we have deleted these three arcs, if we analyze position \fbox{1} we can
delete now arcs \verb!1!, \verb!2! and \verb!4! obtaining Figure 3.

\begin{center}
\epsfxsize=8cm
\epsfbox{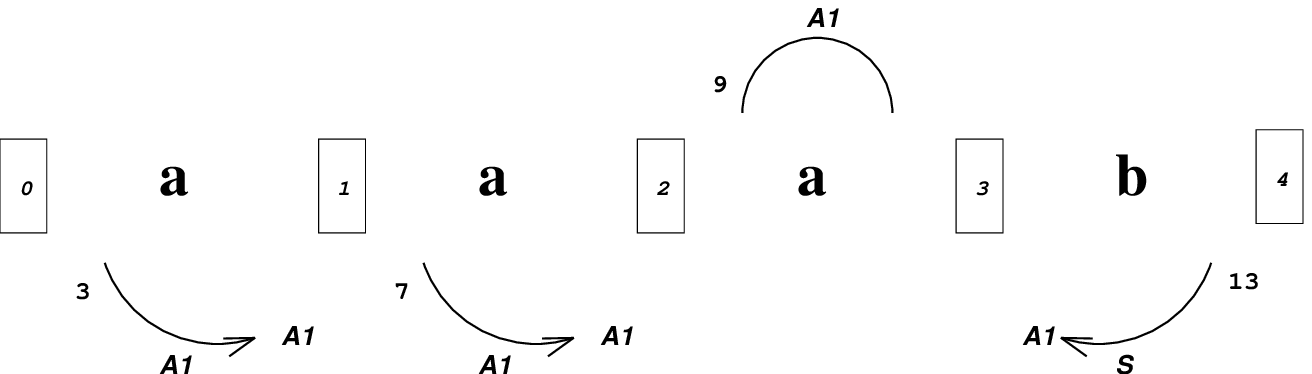}
\end{center}
\begin{center}
\small {\bf Figure 3}
\end{center}

Therefore, our next goal will be to define formally the relations
between arcs that guarantee their success during parsing.

\section{The Mathematical Kernel}

\subsection{Bottom-Up Derivation}

Given a context-free grammar $G = <G_T,G_N,G_P,G_R>$ where we have 
distinguished their
set of productions $G_P$, roots $G_R$, terminal symbols $G_T$,
non-terminal symbols $G_N$ and vocabulary $G_V = G_T \cup G_N$, we will define
the bottom-up derivation as follows.
Let be $\delta \in G_V$ and $\Delta,\Gamma,\Omega \in G_V^*$.
The direct bottom--up derivation in $G$, $\longrightarrow_G$, 
is defined as:
\begin{center}
        $\Gamma\Delta\Omega \longrightarrow_G \Gamma\delta\Omega$
        \siff $\delta \longrightarrow \Delta \in G_P$
\end{center}

The bottom--up derivation in $G$, $\Longrightarrow_G$, will be defined as
the reflexive and transitive closure of the direct bottom--up derivation:

\begin{center}
        $\Gamma \Longrightarrow_G \Omega$ \siff
        $\exists \Delta_1,\ldots,\Delta_n \in G_V^*$
        such that
        $\forall i_{(1 \le i < n)} \Delta_i \longrightarrow_G
        \Delta_{i+1}$\\ where $\Delta_1 \equiv \Gamma$ and 
	$\Delta_n \equiv \Omega$
\end{center}

\subsection{Partial Derivability and Adjacency}

\paragraph{Root Symbols.}
$\alpha$ is a root symbol: 
$R(\alpha)$ \siff $\alpha \in G_R$

\paragraph{Epsilon Symbols.}
$\alpha$ is an epsilon symbol:
$E(\alpha)$ \siff
$\varepsilon \Longrightarrow_G \alpha$  \footnote{$\varepsilon$
is the empty string.}

\paragraph{String of Epsilon Symbols.}
$\Delta$ is a string of epsilon symbols:
        $E(\Delta)$ \siff
        $\forall \delta \in \Delta (E(\delta))$

\paragraph{Left Partial Derivability.}
$\beta$ is a left partial derivation of $\alpha$:
\begin{center}
        $\alpha \longmapsto^*_l \beta$ \siff
        $\exists \Gamma, \Delta, \Omega \in G_V^*$ such that
        $(\Gamma\alpha\Delta \Longrightarrow_G \Gamma\beta\Omega)$.
\end{center}
We define $LPD(\alpha) = \{\beta \in G_V :
\alpha \longmapsto^*_l \beta\} \cup \{\alpha\}$
\footnote{We will consider that a symbol is a left partial
derivation of itself.}

\paragraph{Right Partial Derivability.}
$\beta$ is a right partial derivation of $\alpha$:
\begin{center}
        $\alpha \longmapsto^*_r \beta$ \siff
        $\exists \Gamma, \Delta, \Omega \in G_V^*$ such that
        $(\Gamma\alpha\Delta \Longrightarrow_G \Omega\beta\Delta)$
\end{center}
We define $RPD(\alpha) = \{\beta \in G_V :
\alpha \longmapsto^*_r \beta\} \cup \{\alpha\}$
\footnote{We will consider that a symbol is a right partial
derivation of itself.}

\paragraph{Primary Adjacency.}
$\beta$ is a primary adjacent of $\alpha$:
\begin{center}
        $\alpha \Uparrow \beta$ \siff
        $\exists \delta \in G_V$ and $\exists \Gamma, \Omega, \Delta
        \in G_V^*$ such that $(\delta \longrightarrow
        \Gamma\alpha\Delta\beta\Omega \in G_P \wedge
        E(\Delta))$
\end{center}

\paragraph{Left Adjacency.}
$\beta$ is a left adjacent of $\alpha$:
\begin{center}
        $\alpha \Uparrow^*_l \beta$ \siff
        $\exists \gamma \in LPD(\alpha)$ and  
        $\exists \delta \in RPD(\beta)$ such that
        $(\delta \Uparrow \gamma)$
\end{center}
We define  $LA(\alpha) = \{\beta \in G_V :
\alpha \Uparrow^*_l \beta \}$.

\paragraph{Right Adjacency.}
$\beta$ is a right adjacent of $\alpha$:
\begin{center}
        $\alpha \Uparrow^*_r \beta$ \siff
        $\exists \gamma \in RPD(\alpha)$ and 
        $\exists \delta \in LPD(\beta)$ such that
        $(\gamma \Uparrow \delta)$
\end{center}
We define  $RA(\alpha) = \{\beta \in G_V :
\alpha \Uparrow^*_r \beta \}$.

\paragraph{Left--Most Symbol.}
$\alpha$ is a left--most symbol:
\begin{center}
        $LM(\alpha)$ \siff $\exists \delta \in G_V$ such that
        $(\alpha \longmapsto^*_l \delta \wedge R(\delta))$
\end{center}

\paragraph{Right--Most Symbol.}
$\alpha$ is a right--most symbol:
\begin{center}
        $RM(\alpha)$ \siff $\exists \delta \in G_V$ such that
        $(\alpha \longmapsto^*_r \delta \wedge R(\delta))$
\end{center}

\subsection{Coverage Tables}

Finally we present the formal definition of the coverage tables
which are in charge of triggering the events of the bidirectional
parser.

For each symbol of a grammar, $\alpha \in G_V$, their left, $LC1(\alpha)$
and $LC2(\alpha)$, medium, $MC(\alpha)$, and right, $RC(\alpha)$, coverages
are defined as sets of productions in the following way:

\begin{center}
$LC1(\alpha) = \{ (\delta \longrightarrow \alpha \in G_P) : 
		\delta \in G_N \}$\\
$LC2(\alpha) = \{ (\delta \longrightarrow \alpha\Omega \in G_P) :
		\delta \in G_N \wedge \Omega \in G_V^* 
			\wedge \neg E(\Omega) \}$ \\
$MC(\alpha) = \{ (\delta \longrightarrow \Delta\alpha\Omega \in G_P) :
		\delta \in G_N \wedge \Omega,\Delta \in G_V^*
			 \wedge \neg E(\Delta) \wedge \neg E(\Omega) \}$ \\
$RC(\alpha) = \{ (\delta \longrightarrow \Delta\alpha \in G_P) :
		\delta \in G_N \wedge \Delta \in G_V^*
			 \wedge \neg E(\Delta) \}$ 
\end{center}

\section{The Parsing Algorithm}

\subsection{Parsing Input.} 
The main task of the lexical analyzer is to separate the input
string in a set of items, each one associated with one or more
(lexical ambiguity) pre-terminal symbols (syntactic categories).
Our parsing algorithm is also able to deal with ``multi-word
expressions'' and ``multi-expression words''\footnote{Words that
contain more than one lexical unit, such as clitics in Spanish
or compounds in German.}.

In any case, the parsing input will be a list of breaking points
and a set of pre-terminal symbols, each one associated with a lexical
unit (a portion of the input string) and two breaking points.
For instance, we can consider the input string {\tt a a a b}.
This string will be lexically analyzed obtaining 5 breaking points
and 4 pre-terminal symbols:

\begin{center}
$\sfbox{0} a \sfbox{1} a \sfbox{2} a \sfbox{3} b \sfbox{4}$
\end{center}

\subsection {Step 1: $CaD$ creation.}

For each breaking point we will generate a $CaD$ (collection and 
diffusion of information) structure, which has 6 fields: the first four
fields are lists of $Events$ and the two last ones are lists of $Nodes$:
{\em tole} (events arriving at the $CaD$ from the right side),
{\em frle} (events going to the left from the $CaD$),
{\em tori} (events arriving at the $CaD$ from the left side),
{\em frri} (events going to the right from the $CaD$),
{\em ndle} (nodes at the left of the $CaD$) and
{\em ndri} (nodes at the right of the $CaD$).

If the lexical analyzer has obtained $n$ breaking points, then we will
store the $CaD$ structures as a matrix of $n$ pointers to $CaD$ structures.
We will call this matrix $CaD_{root}$.

\subsection {Step 2: $Node$ creation.}

For each element 
\verb!<lexical_unit,pre-terminal_symbol,fbp!\footnote{The first or left 
breaking point of the lexical unit.}
\verb!,lbp!\footnote{The last or right breaking point of the lexical unit.}
\verb!>!
we will generate a $Node$ structure, which has the following fields:
{\em grsymbol} (grammar symbol) and
{\em cmanalysis} (complex analysis, a list of $Analysis$ structures).
The new node $newNode$ will be associated with
the corresponding $CaD$ structures:

\begin{center}
$CaD_{root}[fbp]\verb!->!ndri = AddNode(newNode)$\\
$CaD_{root}[lbp]\verb!->!ndle = AddNode(newNode)$
\end{center}

\subsection {Step 3: $Event$ creation.}

For each node created at step 2, we will generate their correspondent events
using the coverage tables. 
An $Event$ has the following fields:
{\em grprod} (production or grammar rule),
{\em leftdot} (left dot),
{\em rightdot} (right dot),
{\em leftlinks} (list of $Link$ structures associated with the
	left extreme),
{\em rightlinks} (list of $Link$ structures associated with the
	right extreme) and
{\em status} (logical status).
Let us suppose that the node created ($newNode$) has been associated with
the grammar symbol $\alpha$. Then:

For each production $p \in LC1(\alpha)$ we will create the appropriate
new event ($newEvent$) and:
\begin{center}
$CaD_{root}[fbp]\verb!->!frri = AddEvent(newEvent)$\\
$CaD_{root}[lbp]\verb!->!frle = AddEvent(newEvent)$
\end{center}

For each production $p \in LC2(\alpha)$ we will create the appropriate
new event ($newEvent$) and:
\begin{center}
$CaD_{root}[fbp]\verb!->!frri = AddEvent(newEvent)$\\
$CaD_{root}[lbp]\verb!->!tole = AddEvent(newEvent)$
\end{center}

For each production $p \in MC(\alpha)$ we will create the appropriate
new event ($newEvent$) and:
\begin{center}
$CaD_{root}[fbp]\verb!->!tori = AddEvent(newEvent)$\\
$CaD_{root}[lbp]\verb!->!tole = AddEvent(newEvent)$
\end{center}

For each production $p \in RC(\alpha)$ we will create the appropriate
new event ($newEvent$) and:
\begin{center}
$CaD_{root}[fbp]\verb!->!tori = AddEvent(newEvent)$\\
$CaD_{root}[lbp]\verb!->!frle = AddEvent(newEvent)$
\end{center}

\subsection {Step 4: $Link$ creation.}

For each event created we have to analyze their possible links with other
events. This operation is internal to the $CaD$ structure according to
the following criteria:

\begin{center}
\epsfbox{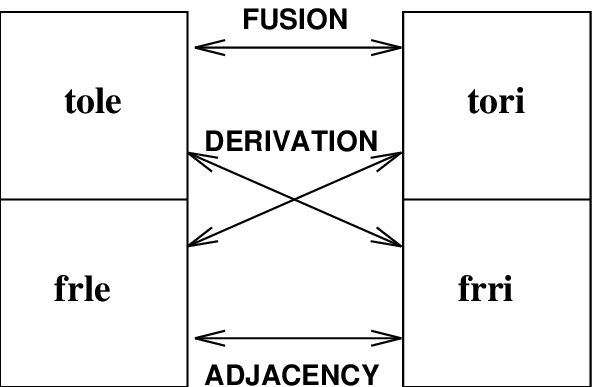}
\end{center}
\begin{center}
\small {\bf Figure 4}. Links inside a $CaD$ structure.
\end{center}

\subsubsection{Analyses of Partial Derivations}

These analyses are applied over the open extremes of an event. Basically,
we have to check if the symbol needed is partially derivable from the
real symbol.

We will distinguish two types of analyses of derivations
depending on the direction of the open extreme of the event.

\paragraph{Left--Derivation Analysis: $TOLE-FRRI$.}

Let us suppose that an event $evtole$ is applying the production
$\delta \longrightarrow \delta_1 \ldots \delta_n$
over the surface
$\Gamma \delta_h \ldots \delta_i \gamma_1 \ldots \gamma_j \Omega$
where $1 \le h \le i < n$ and $1 \le j \le m$. 
In fact, this event is making a prediction
of the (required) symbol $\delta_{i+1}$ over the (real) symbol
$\gamma_1$. The right extreme of this event will be associated with 
the component $tole$ of a $CaD$ structure. In this case, we have to check if
there exists a second event, $evfrri$, which left extreme belongs to the 
$frri$ 
field of the same $CaD$, such that $\delta_{i+1} \in LPD(\gamma)$, 
where $\gamma$ is the left-hand side of the production associated 
with $evfrri$: $\gamma \longrightarrow \gamma_1 \ldots \gamma_m$.

\paragraph{Right--Derivation Analysis: $TORI-FRLE$.}

Let us suppose that an event $evtori$ is applying the production
$\delta \longrightarrow \delta_1 \ldots \delta_n$
over the surface
$\Gamma \gamma_j \ldots \gamma_m \delta_h \ldots \delta_i \Omega$
where $1 < h \le i \le n$ and $1 \le j \le m$. 
In fact, this event is making a prediction
of the (required) symbol $\delta_{h-1}$ over the (real) symbol
$\gamma_m$. The left extreme of this event will be associated 
with the component $tori$ of a $CaD$ structure. In this case, we have to 
check if there exists a second event, $evfrle$, which right extreme 
belongs to the $frle$ field of the same $CaD$, such that 
$\delta_{h-1} \in RPD(\gamma)$,
where $\gamma$ is the left-hand side of the production associated
with $evfrle$: $\gamma \longrightarrow \gamma_1 \ldots \gamma_m$.

\subsubsection{Analyses of Adjacencies}

These analyses are applied over the closed extremes of an event. Basically,
we have to check the adjacency relation between the symbol
that the event will generate (the left--hand side of the production)
and the symbol that appears next to the closed extreme of the
event.

We will distinguish two types of analyses of adjacencies
depending on the direction of the closed extreme of the event.

\paragraph{Left--Adjacency Analysis: $FRRI-FRLE$.}

Let us suppose that an event $evfrri$ is applying the production
$\delta \longrightarrow \delta_1 \ldots \delta_n$
over the surface
$\Gamma \gamma_j \ldots \gamma_m \delta_1 \ldots \delta_i \Omega$
where $1 \le i \le n$ and $1 \le j \le m$. 
The left extreme of this event belongs to the $frri$ field of a 
$CaD$ structure. Then we have to analyze if there exists an $evfrle$ 
event which right extreme belongs to the $frle$ field of the
same $CaD$ such that $\gamma \in LA(\delta)$, where $\gamma$ is the
lhs of the production of $evfrle$:
$\gamma \longrightarrow \gamma_1 \ldots \gamma_m$.

If $\Gamma \gamma_j \ldots \gamma_m$ is empty, that is, the $CaD$ 
associated with the left extreme of $evfrri$ is the first one,
then we have to check if $LM(\delta)$.

\paragraph{Right--Adjacency Analysis: $FRLE-FRRI$.}

Let us suppose that an event $evfrle$ is applying the production
$\delta \longrightarrow \delta_1 \ldots \delta_n$
over the surface
$\Gamma \delta_i \ldots \delta_n \gamma_1 \ldots \gamma_j \Delta$
where $1 \le i \le n$ and $1 \le j \le m$. 
The right extreme of this event belongs to the $frle$ field of a 
$CaD$ structure. Then we have to analyze if there exists an $evfrri$ 
event which left extreme belongs to the $frri$ field of the
same $CaD$ such that $\gamma \in RA(\delta)$, where $\gamma$ is the
lhs of the production of $evfrri$:
$\gamma \longrightarrow \gamma_1 \ldots \gamma_m$.

If $\gamma_1 \ldots \gamma_j \Delta$ is empty, that is, the $CaD$ 
associated with the left extreme of $evfrle$ is the last one,
then we have to check if $RM(\delta)$.

\subsubsection{Analyses of Fusions}

\paragraph{Left--Fusion Analysis: $TORI-TOLE$.}

Let us suppose that an event $evtori$ is applying the production
$\delta \longrightarrow \delta_1 \ldots \delta_n$
over the surface 
$\Gamma \gamma \delta_i \ldots \delta_j \Delta$
where $1 < i \le j \le n$.
The left extreme of this event belongs to the $tori$ field of a 
$CaD$ structure. Then we have to analyze if there exists an $evtole$ 
event in the $tole$ field of the same $CaD$ such that $evtole$ is 
applying the same production that $evtori$ over the surface 
$\delta_{h} \ldots \delta_{i-1}$ where $1 \le h$.

\paragraph{Right--Fusion Analysis: $TOLE-TORI$.}

Let us suppose that an event $evtole$ is applying the production
$\delta \longrightarrow \delta_1 \ldots \delta_n$
over the surface 
$\Gamma \delta_i \ldots \delta_j \gamma \Delta$
where $1 \le i \le j < n$.
The right extreme of this event belongs to the $tole$ field of a 
$CaD$ structure. Then we have to analyze if there exists an $evtori$ 
event in the $tori$ field of the same $CaD$ such that $evtori$ is 
applying the same production that $evtole$ over the surface 
$\delta_{j+1} \ldots \delta_k$ where $k \le n$.

\paragraph{$Link$ creation.}

Each time an analysis is successfull, we will generate a $Link$ structure
between the two events involved. For $LM$ and $RM$ analysis the $Link$ will
have only one event.

\subsection {Step 5: Event's Logical Status.}

Each time a $Link$ is created we have to study the logical status of the
events involved. Also, at the end of the analysis of the links of an event
(step 4) we will analyze its logical status.

Let be $e$ an event applying the production
$\delta \longrightarrow \delta_1 \ldots \delta_{i-1} \bullet \delta_i
 \ldots \delta_j \bullet \delta_{j+1} \ldots \delta_n$

\subsubsection{Closed-Closed events ($FRRI+FRLE$): $i=1$ and $j=n$.}

\noindent
\verb#        if ((!e->leftlinks) || (!e->rightlinks))#\\
\verb#            nstatus = DELETE#\\
\verb#        else#\\
\verb#            nstatus = RUN #

\subsubsection{Closed-Open events ($FRRI+TORI$): $i=1$ and $j<n$.}

\noindent
\verb#        if (!e->leftlinks)#\\
\verb#            nstatus = DELETE#\\
\verb#        else if (e->rightlinks)#\\
\verb#            nstatus = DERIVATION#\\
\verb#        else if (#$E(\delta_{j+1})$\verb#)#\\
\verb#            nstatus = EPSILON#\\
\verb#        else#\\
\verb#            nstatus = DELETE#

\subsubsection{Open-Closed events ($TOLE+FRLE$): $i>1$ and $j=n$.}

\noindent
\verb#        if (!e->rightlinks)#\\
\verb#            nstatus = DELETE#\\
\verb#        else if (e->leftlinks)#\\
\verb#            nstatus = DERIVATION#\\
\verb#        else if (#$E(\delta_{i-1})$\verb#)#\\
\verb#            nstatus = EPSILON#\\
\verb#        else#\\
\verb#            nstatus = DELETE#

\subsubsection{Open-Open events ($TOLE+TORI$): $i>1$ and $j<n$.}

\noindent
\verb#        if ((e->leftlinks) && (e->rightlinks))#\\
\verb#            nstatus = DERIVATION#\\
\verb#        else if ((e->leftlinks) && (#$E(\delta_{j+1})$\verb#))#\\
\verb#            nstatus = EPSILON#\\
\verb#        else if ((e->rightlinks) && (#$E(\delta_{i-1})$\verb#))#\\
\verb#            nstatus = EPSILON#\\
\verb#        else#\\
\verb#            nstatus = DELETE#

If \verb!nstatus! is different that \verb!e->status! we will change the
logical status of the event. To improve the efficiency it is possible
to maintain four lists of events (DERIVATION, RUN, DELETE and EPSILON).
To change the status of an event implies to move the event from one list
to another, but this may be done in constant time.

\subsubsection {Step 6: Parsing Cycle.}

This is the kernel of the algorithm:

\noindent
\verb#        cycle = 1#\\
\verb#        while (cycle)#\\
\verb#            cycle = 0#\\
\verb#            if (event = GetEpsilonEvent())#\\
\verb#                cycle = 1#\\
\verb#                EpsilonExpansion(event)#\\
\verb#            else if (event = GetDeleteEvent())#\\
\verb#                cycle = 1#\\
\verb#                DeleteEvent(event)#\\
\verb#            else if (event = GetRunEvent())#\\
\verb#                cycle = 1#\\
\verb#                RunEvent(event)#\\
\verb#            else if (link = GetFusionLink())#\\
\verb#                cycle = 1#\\
\verb#                FusionLink(link)#\\

The functions \verb!Get*! return the first element of the correspondent
list and change the head of the list to the following element, which
are constant operations.

\paragraph{6.1.- Epsilon Expansion.}
This operation moves the left dot one position to the left or the
right dot one position to the right, depending on the open extreme
marked as EPSILON.

\paragraph{6.2.- Delete Event.}
To delete an event implies to delete it and their links.

\paragraph{6.3.- Run Event.}
To run a closed-closed event involves the application of a grammar
rule, incorporating a new node (step 2).
But if this node has been previously created between the same $CaD$
structures, we can obtain a representation model based on subtree-sharing
and local ambiguity packing, associating the analysis correspondent
to the last one with the previously created node.
This way, a node will have a list of $Analysis$ structures, and this structure
is defined as a list of $Node$ structures. The result of this 
mechanism is a representation based on {\em virtual} relations
between the skeleton of the parse forest and the nodes included in it.

\paragraph{6.4.- Fusion Events.}
Let us consider the production
$\delta \longrightarrow \delta_1 \ldots \delta_h \ldots \delta_i \delta_{i+1} 
\ldots \delta_j \ldots \delta_n$
and two events 
$e_1 : \delta \longrightarrow \delta_1 \ldots \bullet 
	\delta_h \ldots \delta_i \bullet \delta_{i+1} \ldots \delta_n$ 
and 
$e_2 : \delta \longrightarrow \delta_1 \ldots \delta_i \bullet
	\delta_{i+1} \ldots \delta_j \bullet \ldots \delta_n$.

If there exists a fusion link ($e_1e_2link$) between $e_1$ (rightlinks)
and $e_2$ (leftlinks) in the context $\delta_i \delta_{i+1}$ the application
of their fusion will generate the following actions:

{\bf Case 6.4.1: Fusion with Double Derivation:}

\noindent
\verb#        if ((#$e_1$\verb#->rightlinks) && (#$e_2$\verb#->leftlinks))#\\
\verb#            Create a new event#
$e_n : \delta \longrightarrow \delta_1 \ldots \bullet \delta_h \ldots
	\delta_i \delta_{i+1} \ldots \delta_j \bullet \ldots \delta_n$ \\

{\bf Case 6.4.2: Fusion with Single Right Derivation:}

\noindent
\verb#        else if (#$e_1$\verb#->rightlinks)#\\
\verb#            Modify #
$e_2 : \delta \longrightarrow \delta_1 \ldots \bullet \delta_h \ldots
	\delta_i \delta_{i+1} \ldots \delta_j \bullet \ldots \delta_n$ \\

{\bf Case 6.4.3: Fusion with Single Left Derivation:}

\noindent
\verb#        else if (#$e_2$\verb#->leftlinks)#\\
\verb#            Modify #
$e_1 : \delta \longrightarrow \delta_1 \ldots \bullet \delta_h \ldots
	\delta_i \delta_{i+1} \ldots \delta_j \bullet \ldots \delta_n$ \\

{\bf Case 6.4.4: Fusion without Derivation:}

\noindent
\verb#        else #\\
\verb#            Modify #
$e_1 : \delta \longrightarrow \delta_1 \ldots \bullet \delta_h \ldots
	\delta_i \delta_{i+1} \ldots \delta_j \bullet \ldots \delta_n$ \\
\verb#            Delete #$e_2$\\

\section{Implementation and Experimental Results}

This algorithm has been implemented in C including a specific layer
for the memory management that improves the classical operations of
{\bf malloc} and {\bf free}.

Our experimental results show that this algorithm fully eliminates  
parsing ambiguity for recursive, local and non-local dependency
constructions. For this kind of phenomena, the experimental results
show a real complexity of the order $O(n log(n))$ where $n$ is the
length of the input string. \footnote{\cite{Quesada:97} contains
a full description of the algorithm as well as a more detailed
analysis of the experiments, including the grammars, string of
words and results.}

Next, we show the predicted model obtained for
each type of grammar. The dependent variable
{\it T} is the time used for the complete analysis (in
seconds) and the factor used, {\it W}, has been the
length of the input string (number of words).

We show the results obtained with a Simple
Lineal Regression Test for two cases. The first one
uses {\it T} as the response and {\it W log(W)} as
the factor. The second one uses  {\it T/W} as the response
and the same factor {\it W log(W)}. In addition, we have
included Pearson Correlation Coefficients for both cases.

\begin{itemize}
\item Recursive Constructions:
\begin{center}
$T = -5.183 + 219E-7 * (W lg(W)); PCC(T,W lg(W)) = 0.999$\\
$T/W = 0.0001 + 46E-13 * (W lg(W)); PCC(T/W,W lg(W)) = 0.974$
\end{center}

\item Local Dependencies
\begin{center}
$T = -17.82 + 352E-7 * (W lg(W)); PCC(T,W lg(W)) = 0.993$\\
$T/W = 0.0002 + 38E-12 * (W lg(W)); PCC(T/W,W lg(W)) = 0.998$
\end{center}

\item Non-local Dependencies
\begin{center}
$T = -1.031 + 851-8 * (W lg(W)); PCC(T,W lg(W)) = 0.998$\\
$T/W = 0.0002 + 14E-12 * (W lg(W)); PCC(T/W,W lg(W)) = 0.998 $
\end{center}

\end{itemize}

\section{Conclusion}

The problem of parsing natural languages must be studied from three
perspectives: computational, linguistic and formal. In this paper
we have presented a general, sound and efficient natural language
parsing algorithm which accomplishes the main requirements of the
three levels.

The computational layer includes a specific memory management model
and a strategy for grammar compilation. This module has been designed
with the goal of efficiency. The linguistic layer is in charge of general
applicability, and includes basically a mecanism for the integration
of the algorithm with unification grammar. Finally, at the formal level,
the mathematical kernel proposed permits the demostration of the
correctness and soundness of the algorithm \cite{Quesada:97}.

This paper has concentrated on the description of the algorithm itself,
describing the data model and the parsing strategy.

\end{document}